\documentclass{article}

    \PassOptionsToPackage{numbers, compress}{natbib}
 \usepackage[preprint]{neurips_2026}

\usepackage[utf8]{inputenc} 
\usepackage[T1]{fontenc}    
\usepackage[pagebackref=true,breaklinks=true,letterpaper=true,colorlinks,bookmarks=false]{hyperref}
\usepackage{url}            
\usepackage{booktabs}       
\usepackage{amsfonts}       
\usepackage{nicefrac}       
\usepackage{microtype}      
\usepackage[table]{xcolor}  
\usepackage{graphicx}
\usepackage{multirow}
\usepackage{amsmath}
\usepackage[ruled,vlined]{algorithm2e}
\usepackage{cleveref}

\newcommand{\etal}{\emph{et al.}}

\definecolor{c2}{HTML}{FBD9BD}
\definecolor{lp}{HTML}{f7f5fc}
\definecolor{c3}{HTML}{fe793d}
\definecolor{c4}{HTML}{eedeb0}
\definecolor{pp}{HTML}{BC7FCD}
\definecolor{bb}{HTML}{CDE8E5}
\definecolor{rouse}{rgb}{0.981,0.961,0.941}
\definecolor{second}{HTML}{00B0F0}
\definecolor{best}{HTML}{FF0000}

\title{PRISM: Rethinking Atmospheric Scattering Reconstruction as a Unified Understanding and Restoration Model for Real-world Dehazing}

\author{Chengyu Fang$^{1,2}$,\,
        Chunming He$^{2}$,\,
        Yuelin Zhang$^{3}$,\, 
        Chubin Chen$^{1}$,\, 
	Chenyang Zhu$^{1}$,\,\\
    \textbf{Hongqiu Wang$^{5}$,\,
	Longxiang Tang$^{4}$,\,
	Xiu Li$^{1,\dagger}$,\,
    Sina Farsiu$^{2,\dagger}$}\\
	$^1$Tsinghua University,~ $^2$Duke University,~ $^3$CUHK,~ $^4$HKUST, ~ $^5$HKUST(GZ)
 \\
 \url{https://github.com/cnyvfang/PRISM}
 }

\begin{document}

\maketitle

\begin{abstract}
Real-world image dehazing (RID) aims to remove haze-induced degradation from real scenes. This task remains challenging due to non-uniform haze distribution, spatially varying color shifts, and the scarcity of paired real hazy-clean data. In PRISM, we propose \textit{Proximal Scattering Atmosphere Reconstruction (PSAR)}, a physically structured framework that jointly reconstructs the clear scene and scattering variables under the atmospheric scattering model, making the restoration process more interpretable in complex real-world conditions. To bridge the synthetic-to-real gap, we design an online non-uniform haze synthesis pipeline and a \textit{Selective Self-Distillation Adaptation (SSDA)} scheme for unpaired real-world scenarios, which enables the model to selectively learn from high-quality perceptual targets while leveraging its intrinsic scattering understanding to audit residual haze and guide self-refinement. Experiments on real-world benchmarks demonstrate that PRISM achieves competitive performance on RID tasks.
\end{abstract}

\begin{figure*}[h]
    \vspace{-2mm}
	\setlength{\abovecaptionskip}{-0.2cm}
	\begin{center}
		\includegraphics[width=\linewidth]{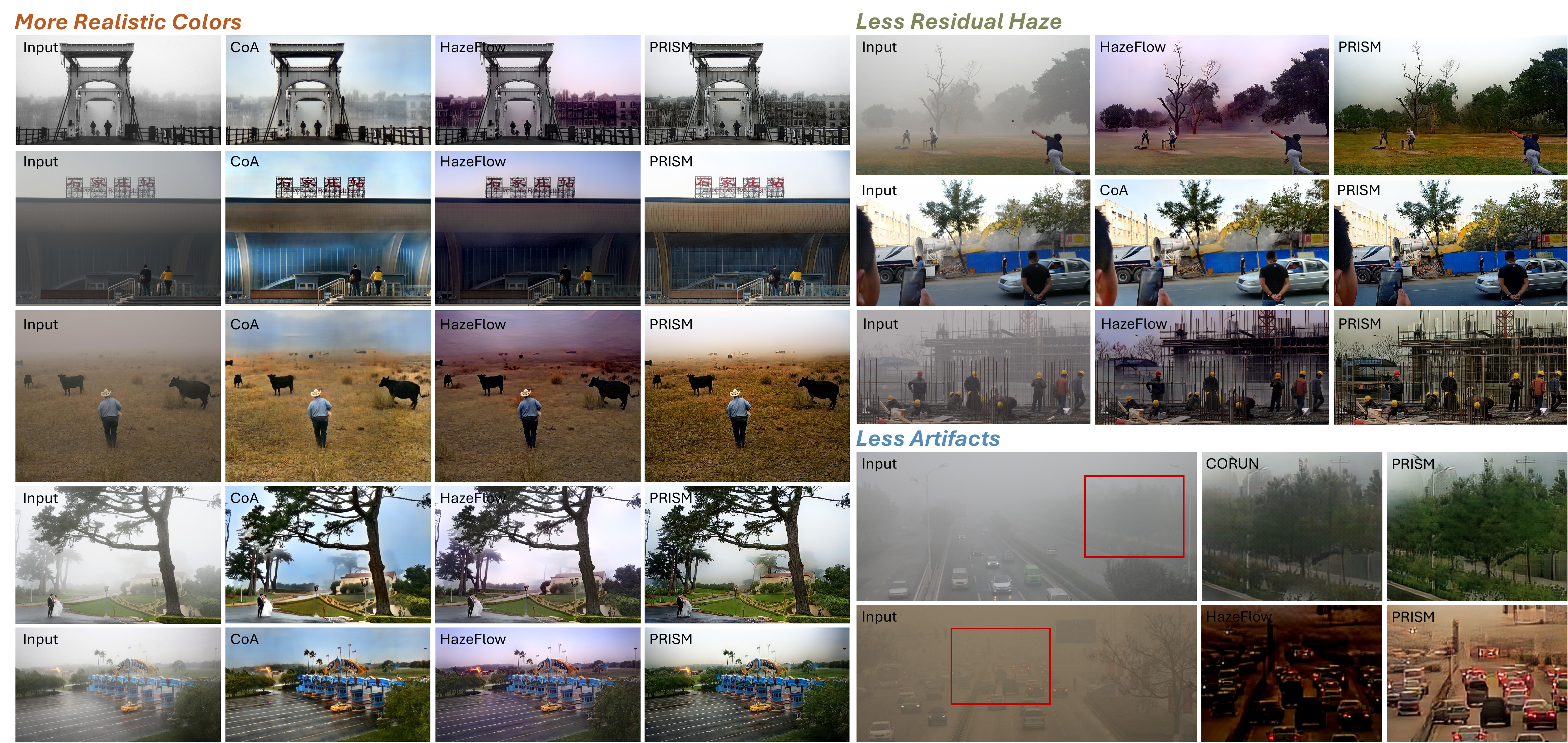}
	\end{center}
	\caption{Our proposed PRISM framework enables effective non-uniform haze removal, producing results with more realistic colors and fewer artifacts than prior methods.}
	\label{fig:head}
    \vspace{-4mm}
\end{figure*}

\section{Introduction}
\label{sec:intro}
Image dehazing is a long-standing problem in intelligent transportation, remote sensing, and general photography. In real-world scenes, haze density is highly non-uniform, and multiple degradations create spatially varying color shifts. These factors make the joint recovery of the clear scene and scattering variables from a single observation highly ill-posed. Many learning-based methods treat dehazing as a direct hazy-to-clear mapping, which often reduces physical interpretability and leads to artifacts in challenging regions. Alternatively, conventional methods based on the atmospheric scattering model (ASM) typically estimate transmission $T$ and atmospheric light $A$, and then invert the model to restore scene radiance $J$. This sequential approach often treats scattering variables as auxiliary byproducts, failing to fully exploit their mutual constraints with the scene content.

To achieve more interpretable restoration, recent works have introduced proximal gradient descent and network unfolding into the ASM framework. For instance, PDN~\cite{yang2018proximal} optimizes the dark channel and transmission map within an unfolded proximal framework. However, its recombination strategy for the final output struggles with the complex composite degradations found in real scenes. Similarly, CORUN~\cite{fang2024real} unfolds gradient steps to jointly update transmission and the scene, improving their coupling. Yet, it assumes a globally constant atmospheric light, which limits its ability to model spatially varying haze and illumination effects in real images.

In real-world scenarios, data distributions pose substantial challenges. Real-world haze is inherently non-homogeneous, often coupled with complex color shifts caused by region-wise color-tinted haze, car lights, and building windows, as well as camera noise and compression artifacts. Existing synthetic haze distributions are typically over-simplified and overly consistent, failing to capture such spatial diversity and stochastic noise. This mismatch, combined with the scarcity of paired hazy-clean images in the wild, creates a synthetic-to-real domain gap that hinders robust adaptation.

To explicitly model haze formation while restoring the scene, we formulate image dehazing as a joint reconstruction problem, where the clear scene and scattering variables are optimized simultaneously. In PRISM, we propose \textbf{Proximal Scattering Atmosphere Reconstruction (PSAR)}, a physically structured framework that performs dehazing through a sequence of optimization-inspired stages. Each stage couples closed-form proximal updates for $J$, $T$, and $A$ with lightweight refinement blocks, transforming the dehazing task into a compact, physics-guided reconstruction process. By representing both $A$ and $T$ as spatial variables, PSAR can better model non-uniform haze and color shifts, instead of forcing these effects to be implicitly absorbed by the restored radiance.

To further improve real-world adaptation, PRISM introduces a \textbf{Selective Self-Distillation Adaptation (SSDA)} scheme together with a lightweight online non-uniform haze synthesis framework. In this framework, an exponential moving average (EMA) teacher produces pseudo-labels, which are evaluated alongside the student's predictions by an ensemble of perceptual metrics. The student distills only from targets that pass a conservative quality gate, reducing its exposure to erroneous pseudo-labels. Beyond this external guidance, the ASM-based model naturally unifies haze modeling and clean-image restoration by estimating transmission and atmospheric light while reconstructing a clean scene.
Leveraging this property, we use the teacher network to audit the student's output for residual haze and guide further refinement. This scattering-based audit complements quality-gated distillation, providing an additional physical constraint for more stable unsupervised adaptation.

\textbf{Our core contributions are threefold:}

(1) We propose PSAR, a physically structured dehazing framework that decouples atmospheric scattering variables from inputs and performs reconstruction through proximal optimization steps.

(2) We introduce an online non-uniform haze synthesis pipeline and SSDA scheme that leverages quality-gated distillation and a physics-guided scattering prior for real-world adaptation.

(3) We evaluate our PRISM framework on real-world image dehazing tasks. Experiments on widely used benchmarks demonstrate that our approach achieves competitive performance.

\section{Related Works}
\label{sec:related_works}

\subsection{Single Image Dehazing} 

Early single image dehazing methods rely on hand-crafted priors. DCP~\cite{he2010single} requires at least one low-intensity channel in local patches, while other priors~\cite{zhu2014single, fattal2014dehazing} use regularities to estimate transmission, but they often fail in sky regions and non-uniform haze. With data-driven learning, deep models have become mainstream~\cite{ren2016single, cai2016dehazenet,li2017aod, liu2019griddehazenet, dong2020multi, feng2024bridging, zhang2026bilevel, liu2019deep} and typically learn direct hazy-to-clear mappings~\cite{wu2021contrastive,qin2020ffa,guo2022image, song2023vision}, yet the lack of physical constraints reduces interpretability and robustness under complex real degradations.
Diffusion models~\cite{fan2026physics, lan2025schrodinger, wang2025learning,lan2025exploiting} have achieved high perceptual quality in dehazing owing to their strong generative capabilities. However, they may occasionally hallucinating unfaithful details, which limiting the credibility issues in downstream tasks. This highlights the necessity of developing physically grounded frameworks that explicitly model the degradation process for reliable reconstruction.

\subsection{Real-world Image Dehazing}

The domain gap between synthetic and real haze remains a key bottleneck for real-world dehazing~\cite{liu2019griddehazenet, dong2020multi, wu2021contrastive, qin2020ffa, guo2022image, ye2022perceiving}. Real haze is entangled with spatially varying degradations, while paired real hazy-clean data are scarce, motivating increasing efforts in real-world settings~\cite{chen2021psd, qiu2023mb, zheng2023curricular, zhang2024depth,wang2024ucl}. To better model real-world haze~\cite{feng2024advancing}, PDN~\cite{yang2018proximal} and CORUN~\cite{fang2024real} introduce explicit physical modeling, but incomplete atmospheric scattering modeling still limits their color fidelity and robustness.
Other studies narrow the gap by improving haze synthesis realism, such as RIDCP~\cite{wu2023ridcp} and Wang~\etal~\cite{WANGcompensated}. HazeFlow~\cite{shin2025hazeflow} further improves realism but requires an additional non-uniform haze field synthesis step, leading to a trade-off between process complexity and haze authenticity. Domain adaptation methods improve generalization~\cite{chen2022disentaglement,yang2022d4,shao2020domain,li2022physically,tsai2025phatnet, fang2025guided}, but adversarial training may introduce artifacts and instability~\cite{goodfellow2014generative}. Some methods combine synthetic and real data~\cite{chen2021psd, cong2024semisupervised, tong2022semiuformer}, yet noisy pseudo-labels in self-distillation may accumulate and degrade restoration quality. Although Colabator~\cite{fang2024real} reduces this issue by up-weighting high-quality regions, it still relies on noisy pseudo-labels and a relatively complex training pipeline. Therefore, an unsupervised adaptation framework is still needed to reduce pseudo-label error propagation while using explicit physical constraints to bridge the synthetic-to-real domain gap.

\section{Methodology}
\label{sec:method}

\subsection{PSAR: Proximal Scattering Atmosphere Reconstruction}
\label{subsec:psar}

We build our model directly on the atmospheric scattering equation:
\begin{equation}
P(x) = T(x)\,J(x) + (1 - T(x))\,A(x), \label{eq:asm-psar}
\end{equation}
where $P(x),J(x)\in\mathbb{R}^{3\times 1}$ are the hazy and clear RGB column vectors at pixel $x$, $A(x)\in\mathbb{R}^{3\times 1}$ is the atmospheric light field, and $T(x)\in\mathbb{R}$ is the scalar transmission.
PSAR treats single image dehazing as the joint reconstruction of $(J,T,A)$ under Eq.~\eqref{eq:asm-psar}.
At each stage $k$, the network starts from coupled proximal updates induced by the same scattering model and then applies lightweight learnable corrections, so that each stage remains explicitly tied to the same physical objective.

\begin{figure*}[t]
	\centering
	\setlength{\abovecaptionskip}{-0.2cm}
	\begin{center}
		\includegraphics[width=\linewidth]{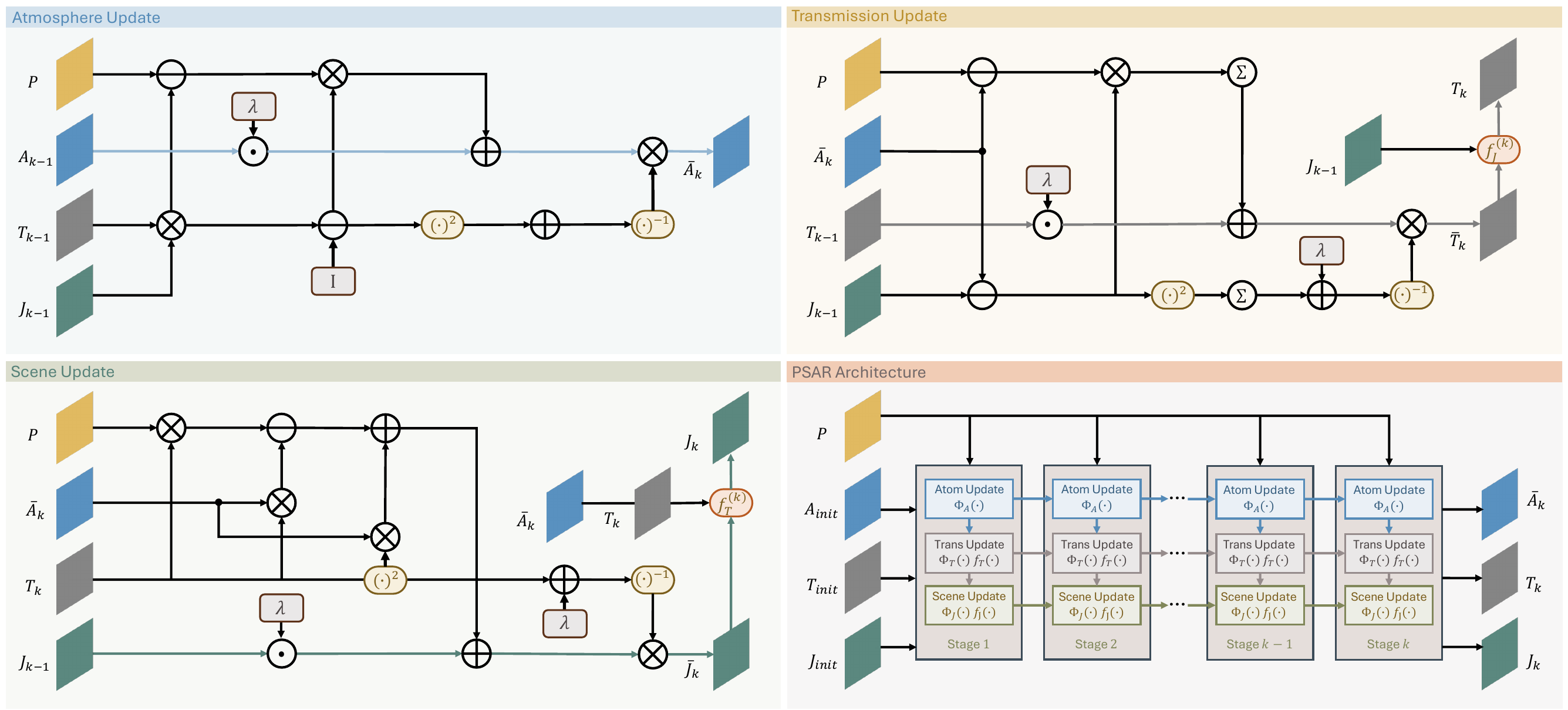}
	\end{center}
 \vspace{1.5mm}
	\caption{Architecture of PSAR at the $k$-th stage. Given the previous estimates $(J_{k-1},T_{k-1},A_{k-1})$, PSAR computes proximal updates for atmospheric light, transmission, and radiance under the same ASM objective, followed by lightweight refinements for $T$ and $J$.}
	\label{fig:Arch}
	\vspace{-0.5cm}
\end{figure*}

\noindent\textbf{Stage-wise scattering objective.}
We measure the mismatch to Eq.~\eqref{eq:asm-psar} by the quadratic data term:
\begin{equation}
    \mathcal{D}(J,T,A)
    =
    \frac{1}{2}
    \sum_{x}
    \|
        P(x) - T(x)\,J(x) - (1-T(x))A(x)
    \|_2^2,
    \label{eq:psar-data-term}
\end{equation}
where $\|\cdot\|_2$ is the Euclidean norm in RGB space.
At stage $k$, PSAR first computes the atmospheric light and transmission proximal points, keeps the atmospheric light branch purely proximal with $A_k=\bar{A}_k$, refines the transmission as $T_k=\bar{T}_k + f_T^{(k)}(\bar{T}_k, J_{k-1})$, and then updates the clear-scene proximal point under $(T_k,A_k)$:
\begin{equation}
\label{eq:psar-prox-triple}
\begin{aligned}
    \bar{A}_k
    &= \operatorname*{arg\,min}_A\;
       \mathcal{D}(J_{k-1},T_{k-1},A)
       + \frac{\lambda_A}{2}\,\|A - A_{k-1}\|_2^2,\\
    \bar{T}_k
    &= \operatorname*{arg\,min}_T\;
       \mathcal{D}(J_{k-1},T,\bar{A}_k)
       + \frac{\lambda_T}{2}\,\|T - T_{k-1}\|_2^2,\\
    \bar{J}_k
    &= \operatorname*{arg\,min}_J\;
       \mathcal{D}(J,T_k,A_k)
       + \frac{\lambda_J}{2}\,\|J - J_{k-1}\|_2^2,
\end{aligned}
\end{equation}
where $\lambda_A,\lambda_T,\lambda_J>0$ are learnable proximal weights.
The quadratic terms serve as stage-to-stage trust regions around $(J_{k-1},T_{k-1},A_{k-1})$.
We use bars to denote the analytic proximal solutions $(\bar{A}_k,\bar{T}_k,\bar{J}_k)$ of these subproblems.
The propagated variables are then refined as
$T_k=\bar{T}_k+f_T^{(k)}(\bar{T}_k,J_{k-1})$ and
$J_k=\bar{J}_k+f_J^{(k)}(\bar{J}_k,T_k,A_k)$, and $(T_k,J_k)$ denote the refined variables passed to the next stage.

We keep $A_k=\bar{A}_k$ without learnable refinement.
Atmospheric light mainly captures low-frequency atmospheric veil and color bias, and its update is a closed-form proximal step anchored to the previous-stage estimate.
Adding a learnable refinement to $A$ would increase parameters in a weakly constrained branch and may amplify errors inherited from the current $J_{k-1}$ and $T_{k-1}$ estimates.
We therefore reserve learnable refinements for $T(x)$ and $J(x)$, where residual structural and radiance errors are more pronounced, and keep the $A$ branch as a bounded proximal update.

\noindent\textbf{Atmospheric light proximal update.}
The first step updates the atmospheric light while keeping $J_{k-1}$ and $T_{k-1}$ fixed.
Most classical formulations use a global atmospheric light $A$, which works under near-uniform color shifts and haze. 
In real scenes, spatially varying color shifts and non-uniform scattering make $A$ spatially varying, so a single RGB vector can force illumination mismatch into $T(x)$ or $J(x)$.
We therefore model atmospheric light as a field $A(x)\in\mathbb{R}^3$ at all stages, enabling locally adaptive updates under the same scattering constraint.
For each pixel $x$, the atmospheric light proximal problem in Eq.~\eqref{eq:psar-prox-triple} becomes:
\begin{equation}
\begin{aligned}
\label{eq:psar-A-problem}
\bar{A}_k(x)
=
\operatorname*{arg\,min}_{A(x)\in\mathbb{R}^3}~
&\frac{1}{2}
\|
    J_{k-1}(x)\,T_{k-1}(x)
    + (1-T_{k-1}(x))A(x)
    - P(x)
\|_2^2\\
&+ \frac{\lambda_A}{2}\,\|A(x) - A_{k-1}(x)\|_2^2,
\end{aligned}
\end{equation}
which is a quadratic least-squares problem in $\mathbb{R}^3$.
Solving the normal equation yields the update:
\begin{equation}
\label{eq:psar-A-update}
    \bar{A}_k(x)
    =
    \frac{
        (1 - T_{k-1}(x))
        (P(x) - J_{k-1}(x)\,T_{k-1}(x))
        + \lambda_A\,A_{k-1}(x)
    }{
        (1 - T_{k-1}(x))^2
        + \lambda_A
    }.
\end{equation}
Here $(1-T_{k-1}(x))$ modulates the contribution of the observation term, while the denominator is at least $\lambda_A>0$, making the update bounded and anchored to $A_{k-1}(x)$.

\noindent\textbf{Transmission proximal update.}
With $J_{k-1}$ and $A_k$ fixed, the scattering model becomes quadratic in $T(x)$, so the update focuses on estimating how much scene radiance is preserved at each pixel under the current atmospheric light. For each pixel $x$, the transmission update from Eq.~\eqref{eq:psar-prox-triple} reads:
\begin{equation}
\begin{aligned}
\label{eq:psar-T-problem}
\bar{T}_k(x)
=
\operatorname*{arg\,min}_{T(x)\in\mathbb{R}}~
&\frac{1}{2}
\|
    T(x)\,J_{k-1}(x)
    + (1-T(x))A_k(x)
    - P(x)
\|_2^2 
+ \frac{\lambda_T}{2}\,\|T(x) - T_{k-1}(x)\|_2^2.
\end{aligned}
\end{equation}
This is a scalar quadratic in $T(x)$.
Writing explicitly over the three RGB channels $c\in\{1,2,3\}$, the analytic solution is:
\begin{equation}
\label{eq:psar-T-update}
    \bar{T}_k(x)
    =
    \frac{
        \lambda_T\,T_{k-1}(x)
        + \displaystyle\sum_{c=1}^3
        (
            A_k^{(c)}(x) - J_{k-1}^{(c)}(x)
        )
        (
            A_k^{(c)}(x) - P^{(c)}(x)
        )
    }{
        \lambda_T
        + \displaystyle\sum_{c=1}^3
        (
            A_k^{(c)}(x) - J_{k-1}^{(c)}(x)
        )^2
    }.
\end{equation}
Here $P^{(c)}(x)$, $J_{k-1}^{(c)}(x)$ and $A_k^{(c)}(x)$ denote the $c$-th components of the corresponding RGB vectors.
The numerator combines the current scattering residual with the previous transmission $T_{k-1}(x)$, and the denominator normalizes by the local contrast between $A_k(x)$ and $J_{k-1}(x)$, yielding a closed-form, data-dependent transmission update that remains anchored to the previous stage.

\noindent\textbf{Clear-scene proximal update.}
After refining the transmission, the third step updates the clear image while keeping $T_k$ and $A_k$ fixed.
With transmission and atmospheric light fixed, the scattering model becomes linear in $J(x)$, so we can solve for the radiance by a closed-form least-squares update under the same physical constraint. For each pixel $x$, the clear-scene update from Eq.~\eqref{eq:psar-prox-triple} becomes:
\begin{equation}
\begin{aligned}
\label{eq:psar-J-problem}
\bar{J}_k(x)
=
\operatorname*{arg\,min}_{J(x)\in\mathbb{R}^3}~
&\frac{1}{2}
\|
    {T}_k(x)\,J(x)
    + (1-{T}_k(x))A_k(x)
    - P(x)
\|_2^2
+ \frac{\lambda_J}{2}\,\|J(x) - J_{k-1}(x)\|_2^2.
\end{aligned}
\end{equation}
This is a quadratic least-squares problem in $\mathbb{R}^3$.
Since $T_k(x)$ is scalar, all three channels share the same scalar coefficients. Solving the normal equation yields:
\begin{equation}
\label{eq:psar-J-update}
    \bar{J}_k(x)
    =
    \frac{
        T_k(x)\,P(x)
        + T_k(x)^2\,A_k(x)
        - T_k(x)\,A_k(x)
        + \lambda_J\,J_{k-1}(x)
    }{
        T_k(x)^2 + \lambda_J},
\end{equation}
where the scalar denominator $T_k(x)^2 + \lambda_J$ divides each RGB component of the numerator.
This update balances adherence to the scattering model with temporal consistency of the clear estimate across stages.

\noindent\textbf{From proximal updates to PSAR stages.}
The three proximal operators in Eqs.~\eqref{eq:psar-A-update}, \eqref{eq:psar-T-update} and \eqref{eq:psar-J-update} define an optimization-inspired backbone.
One PSAR stage is summarized as:
\begin{equation}
\label{eq:psar-stage-compact}
\begin{cases}
\bar{A}_k = \Phi_A(P,J_{k-1},T_{k-1},{A}_{k-1}),\quad A_k = \bar{A}_k\\
\bar{T}_k = \Phi_T(P,J_{k-1},T_{k-1},{A}_k),\quad T_k = \bar{T}_k + f_T^{(k)}(\bar{T}_k, J_{k-1}),\\
\bar{J}_k = \Phi_J(P,J_{k-1},T_k,{A}_k),\quad J_k = \bar{J}_k + f_J^{(k)}(\bar{J}_k, T_k, {A}_k),\\
\end{cases}
\end{equation}
where $\Phi_A,\Phi_T,\Phi_J$ denote the closed-form proximal maps induced by Eq.~\eqref{eq:psar-data-term}, and $f_T^{(k)},f_J^{(k)}$ are shallow learnable refinements.
Stacking several such stages yields a deep architecture in which each stage is built from explicit proximal updates for the same scattering objective, while the lightweight corrections allow the model to adapt to the statistics of real images. The detailed algebraic derivations from Eqs.~\eqref{eq:psar-A-problem}, \eqref{eq:psar-T-problem}, \eqref{eq:psar-J-problem} to Eqs.~\eqref{eq:psar-A-update}, \eqref{eq:psar-T-update}, \eqref{eq:psar-J-update} are deferred to the appendix.

\subsection{Stage 1: Pretraining with Online Non-Uniform Haze Synthesis}

We pretrain the network on synthetic hazy images generated by a depth-guided online scattering model (Alg.~\ref{alg:online-haze-synthesis}), inspired by RIDCP~\cite{wu2023ridcp} and extended to non-uniform density fields and controllable near-camera haze. The design aims to approximate depth-dependent and spatially varying haze observed in real scenes, rather than restricting the model to simple uniform synthetic haze.

\paragraph{Online non-uniform haze synthesis.}
We synthesize diverse haze patterns online without precomputation, as summarized in Alg.~\ref{alg:online-haze-synthesis}. 
Given a clean image $J_{gt}$ and normalized depth $D$, the algorithm samples a base density $\beta_{\mathrm{init}}$ and, with probability $p$, adds a non-negative low-frequency perturbation:
\begin{equation}
\Delta\beta =
\mathcal{S}\!\left(
G_{\sigma_1} * \mathcal{U}(G_{\sigma_0} * z_0)
\right),
\label{eq:lowfreq_field}
\end{equation}
where $\mathcal{U}(\cdot)$ denotes upsampling and $\mathcal{S}(\cdot)$ is a monotone affine rescaling. 
This gives $\beta(x)=\beta_{\mathrm{init}}+\Delta\beta(x)$ for non-uniform haze and $\beta(x)=\beta_{\mathrm{init}}$ for uniform haze, exposing the model to both heterogeneous and homogeneous degradations.

To control foreground haze, we sample $h_{\mathrm{near}}$ and convert it into a depth offset $d_0=-\log(1-h_{\mathrm{near}})/(\beta_{\mathrm{init}}+\varepsilon)$, which is then used with $\beta(x)$ and the normalized depth $D$ to compute $T_{gt}$. 
This allows the density field to vary spatially while keeping the near-camera haze strength controllable. After obtaining $T_{gt}$, we sample the atmospheric light $A_{\mathrm{rend}}$ and combine it with $J_{gt}$ to synthesize the hazy image $P$ by \Cref{eq:asm-psar}, followed by random compression to mimic real image degradation.

\begin{algorithm}[h]
\caption{Online Non-Uniform Haze Synthesis}
\label{alg:online-haze-synthesis}
\KwIn{Clear image $J_{gt}$, normalized depth $D$, probability $p$, near-haze range $[h_{\min}, h_{\max}]$}
\KwOut{Hazy image $P$, transmission $T_{gt}$, atmospheric light offset $A_{\mathrm{rend}}$}

$J_{gt} \leftarrow \mathrm{RandAddGaussianNoise}(\mathrm{RandAdjustLuminance}(J_{gt}))$,\quad
$\beta_{\mathrm{init}} \sim \mathcal{U}(\beta_{\min}, \beta_{\max})$\;

\eIf{\text{with probability } $p$}{
    Generate low-resolution noise $z_0 \sim \mathcal{N}(0, 1)$,\\
    $\beta(x) \leftarrow \beta_{\mathrm{init}} + \mathcal{S}\!\left( G_{\sigma_1} * \mathcal{U}(G_{\sigma_0} * z_0) \right)$ \tcp*{Non-uniform haze}
}{
    $\beta(x) \leftarrow \beta_{\mathrm{init}}$ \tcp*{Uniform haze}
}

$h_{\text{near}} \sim \mathcal{U}(h_{\min}, h_{\max})$,\quad
$d_0 \leftarrow -\log\!(1-h_{\text{near}})/(\beta_{\mathrm{init}} + \varepsilon)$ \tcp*{Foreground haze}
$T_{gt}(x) \leftarrow \exp\!(-\,\beta(x)\,[(1-D(x)) + d_0])$\;

$A_{\mathrm{rend}} \leftarrow \mathrm{Clip}\!\left( \mathcal{U}(A_{\min}, A_{\max})\in\mathbb{R}^1 + \mathcal{U}(-\delta, \delta)\in\mathbb{R}^3, \, 0, \, 1 \right)$\;
$P \leftarrow \mathrm{RandCompress}(T_{gt} \odot J_{gt} + (1 - T_{gt}) \odot A_{\mathrm{rend}})$ \tcp*{Haze rendering}

\Return{$P, T_{gt}, A_{\mathrm{rend}}$}\;
\end{algorithm}

\paragraph{Optimization.}
We train PSAR with an objective that follows its physical decomposition of haze formation. 
The restored radiance $J$ and transmission $T$ are supervised by their targets, while atmospheric light is regularized by a smoothness prior rather than direct pixel-wise supervision:
\begin{equation}
\mathcal{L}_{\text{total}}
=\ \|J-J_{gt}\|_1+\|T-T_{gt}\|_1
+\lambda_A^{\text{tv}}\sum_x\|\nabla A(x)\|_1 
+\lambda_c\frac{d(J,J_{gt})}{d(J,P) + \epsilon}
+\lambda_{\text{adv}}\mathcal{L}_{\text{adv}}(J),
\label{eq:loss_total}
\end{equation}
where $d(\cdot,\cdot)$ denotes the LPIPS distance~\cite{zhang2018unreasonable}, and $\epsilon$ is a small constant. We do not impose direct $\ell_1$ supervision on $A_{\text{rend}}$. During synthesis, $A_{\text{rend}}$ only serves as an image-level anchor for haze rendering, while $J_{gt}$ already contains spatial illumination variations. Matching $A(x)$ to this anchor would bias the atmospheric light branch toward flat fields and limit its adaptation to spatial illumination. Instead, supervision on $J$ and $T$ anchors the scattering reconstruction, while the closed-form $A$ update and total variation prior encourage a smooth yet spatially varying atmospheric light field and prevent $A$ from absorbing high-frequency textures. Beyond pixel-level supervision, the contrastive LPIPS term increases the perceptual separation between the restored radiance $J$ and hazy input $P$ while aligning $J$ with $J_{gt}$. Since LPIPS gradients may also strengthen structured high-frequency artifacts, the adversarial objective encourages natural-image statistics and helps suppress these artifacts.

\subsection{Stage 2: Selective Self-Distillation Adaptation}
\label{subsec:stage2_ssda}

\begin{figure*}[t]
	\centering
	\setlength{\abovecaptionskip}{-0.1cm}
	\begin{center}
		\includegraphics[width=\linewidth]{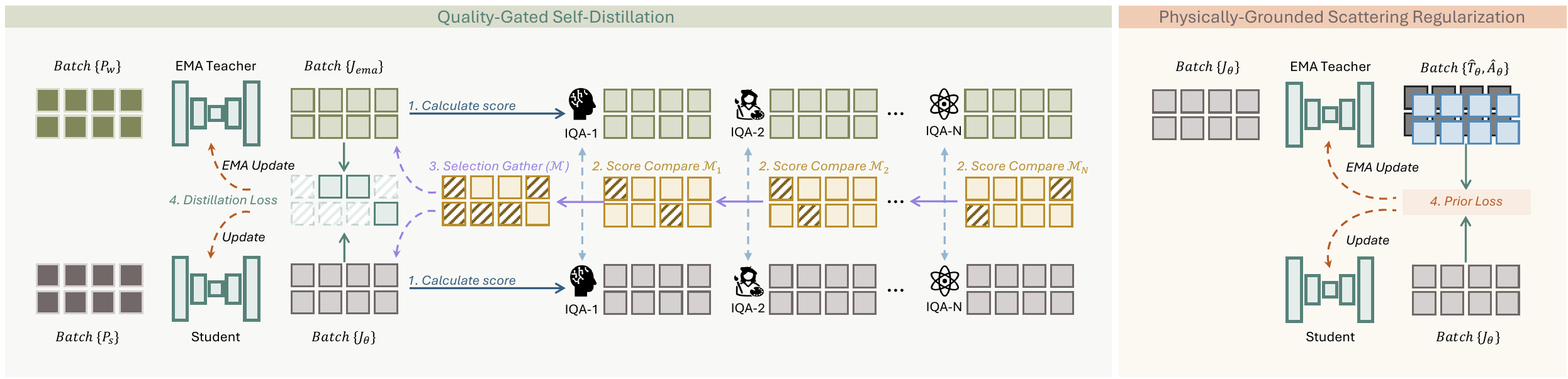}
	\end{center}
	\caption{Pipeline of Selective Self-Distillation Adaptation (SSDA). QGSD selects teacher pseudo-labels only when they outperform the student under all selected perceptual metrics, while PGSR uses the teacher to audit residual haze in the student's output through estimated scattering variables.}
	\label{fig:Pipe}
	\vspace{-0.4cm}
\end{figure*}

Models pretrained on synthetic data often degrade in real-world scenarios due to domain shifts. Since blind unsupervised adaptation can collapse into degenerate solutions or amplify artifacts, we propose the Selective Self-Distillation Adaptation (SSDA) framework to bridge this domain gap (\Cref{fig:Pipe}). Built upon a mean-teacher architecture, SSDA processes a weakly augmented real hazy image $P_w$ through the EMA teacher $\mathcal{F}_{\text{ema}}$ to generate a stable pseudo-label $J_{\text{ema}}$, which guides the student $\mathcal{F}_{\theta}$ operating on a strongly augmented view $P_s$. To ensure robust domain adaptation, SSDA uses two complementary constraints: Quality-Gated Self-Distillation (QGSD) for external perceptual guidance, and a Physically-Grounded Scattering Regularization (PGSR) for internal physical consistency.

\paragraph{Quality-Gated Self-Distillation (QGSD).}
Directly distilling all teacher predictions may propagate unreliable pseudo-labels on out-of-distribution data. To reduce dependence on any single filtering criterion, we introduce a strict filtering mechanism guided by an ensemble of No-Reference Image Quality Assessment (NR-IQA) metrics, denoted as $\Omega = \{\mathcal{Q}_i\}_{i=1}^N$. This module authorizes distillation if and only if the teacher's prediction strictly outperforms the student's across all selected criteria. Specifically, we define a binary update mask $\mathcal{M}$ for each sample:
\begin{equation}
    \mathcal{M} = \prod_{i=1}^N \mathbb{I}\left( \mathcal{Q}_i(J_{\text{ema}}) \succ \mathcal{Q}_i(J_{\theta}) \right),
    \label{eq:update_mask}
\end{equation}
where $\mathbb{I}(\cdot)$ is the indicator function. In our implementation, $\Omega = \{\text{MUSIQ}, \text{NIMA}\}$~\cite{ke2021musiq,talebi2018nima} serves as a comprehensive perceptual safeguard, evaluating both multi-scale visual quality and high-level aesthetics. Gated by $\mathcal{M}$, our selective distillation objective integrates a pixel-wise loss and a contrastive perceptual loss:
\begin{equation}
    \mathcal{L}_{\text{distill}} = \mathcal{M} \cdot ( \| J_{\theta} - \text{sg}(J_{\text{ema}}) \|_1 + \lambda_{c} \frac{d(J_{\theta}, \text{sg}(J_{\text{ema}}))}{d(J_{\theta}, P_{s}) + \epsilon}),
    \label{eq:loss_distill}
\end{equation}
where $\text{sg}(\cdot)$ denotes the stop-gradient operation, and $d(\cdot, \cdot)$ computes the LPIPS distance. By restricting updates to pseudo-labels that pass all quality criteria, this intersection logic reduces the risk of unreliable supervision during adaptation. Because QGSD relies entirely on final dehazed outputs, it functions as a model-agnostic module for robust real-world adaptation.

\paragraph{Physically-Grounded Scattering Regularization (PGSR).}
While quality-gated distillation provides reliable external perceptual guidance, it does not inherently enforce internal physical consistency. Since transmission and atmospheric light estimation encode explicit scene-dependent scattering cues, they provide physically meaningful constraints for haze-free image reconstruction. To exploit this, we introduce a self-reinforced scattering prior. We utilize the stable teacher network $\mathcal{F}_{\text{ema}}$ to physically audit the student's dehazed prediction $J_{\theta}$ for any residual haze:
\begin{equation}
    (\hat{J}_{\text{ema}}, \hat{T}_{\text{ema}}, \hat{A}_{\text{ema}}) = \mathcal{F}_{\text{ema}}(J_{\theta}).
    \label{eq:prior_pipeline}
\end{equation}
If $J_{\theta}$ is completely haze-free, the estimated transmission $\hat{T}_{\text{ema}}$ should ideally approach $\mathbf{1}$. However, uniformly penalizing $\hat{T}_{\text{ema}}$ often induces artifacts in smooth or overexposed regions, where the scattering prior is less reliable. To address this, we construct a spatially aware brightness-and-atmospheric light weight map, $W_{\text{BAW}}$, for reliable physical auditing:
\begin{equation}
    W_{\text{BAW}} = \max(W_{\text{dist}}, W_{\text{tex}}) \odot W_{\text{dist}} \odot W_{\text{high}},
    \label{eq:weight_baw}
\end{equation}
Here, $W_{\text{dist}}$ serves as the primary reliability gate by favoring pixels sufficiently separated from the estimated atmospheric light. 
$W_{\text{tex}}$ enhances textured regions, while $W_{\text{high}}$ suppresses bright or near-saturated pixels. 
These three terms are defined as:
\begin{equation}
    W_{\text{dist}} = \mathcal{S}(\frac{1}{3}\sum_{c} \left|J_{\theta}^c - \hat{A}_{\text{ema}}^c\right|), \quad
    W_{\text{tex}}  = \mathcal{S}(\|\nabla \mathrm{Gray}(J_{\theta})\|), \quad
    W_{\text{high}} = 1 - \mathcal{S}(\max_{c} J_{\theta}^c).
\label{eq:baw_subweights}
\end{equation}
In this way, $W_{\text{BAW}}$ uses atmospheric light deviation as the main condition, texture as an auxiliary cue, and highlight masking as a final exclusion term.

Guided by $W_{\text{BAW}}$, we further apply a robust directional hinge loss to $\hat{T}_{\text{ema}}$. Instead of rigidly forcing the transmission toward $\mathbf{1}$, which may over-suppress natural atmospheric perspective and introduce visual artifacts, we adopt a relaxed target $T_{\text{target}} = 0.9$:
\begin{equation}
    \mathcal{L}_{\text{prior}} = \frac{1}{\|W_{\text{BAW}}\|_1 + \epsilon} \sum ( W_{\text{BAW}} \odot \max(0, T_{\text{target}} - \hat{T}_{\text{ema}})).
    \label{eq:loss_prior}
\end{equation}
By penalizing underestimated transmission only in reliable, non-saturated, atmospheric-light-separated regions, this prior encourages physically consistent scenes with fewer artifacts.

\paragraph{Overall Adaptation Objective.}
The complete objective for the SSDA stage seamlessly integrates the QGSD, the PGSR, and the atmospheric total variation regularization defined in Stage 1:
\begin{equation}
    \mathcal{L}_{\text{adapt}} = \mathcal{L}_{\text{distill}} + \mathcal{L}_{\text{prior}} + \lambda_A^{\mathrm{tv}}\sum_x\|\nabla A(x)\|_1.
    \label{eq:loss_adapt}
\end{equation}
During adaptation, the student updates the EMA teacher, and the teacher provides quality-gated pseudo-labels and scattering-based residual-haze audits for subsequent iterations.
This controlled loop is designed to improve real-world adaptation while limiting updates from low-quality pseudo-labels.

\section{Experiments}
\label{sec:experiments}

\begin{table*}[t]
\centering
\setlength{\abovecaptionskip}{0cm}
\caption{Quantitative results on RTTS~\cite{li2019benchmarking} and Fattal's~\cite{fattal2014dehazing} datasets. The best result of each metric is marked in \textbf{\textcolor{best}{red}}, and the second-best result is marked in \textbf{\textcolor{second}{blue}}.}
\label{tab:rtts_fattal}
\setlength{\tabcolsep}{2.5mm}
\resizebox{\textwidth}{!}{
\begin{tabular}{l|c|ccc|ccc}
\toprule
\multirow{2}{*}{Method} & \multirow{2}{*}{Venue}
& \multicolumn{3}{c|}{RTTS~\cite{li2019benchmarking}}
& \multicolumn{3}{c}{Fattal's~\cite{fattal2014dehazing}} \\
\cline{3-8}
& &
{\cellcolor{gray!20}FADE$\downarrow$}
& {\cellcolor{gray!20}PAQ2PIQ$\uparrow$}
& {\cellcolor{gray!20}CLIPIQA$\uparrow$}
& {\cellcolor{gray!20}FADE$\downarrow$}
& {\cellcolor{gray!20}PAQ2PIQ$\uparrow$}
& {\cellcolor{gray!20}CLIPIQA$\uparrow$} \\
\midrule
Hazy Inputs & --- 
& 2.484 & 66.05 & 0.388
& 1.061 & 71.54 & 0.506 \\

PDN~\cite{yang2018proximal} & ECCV 2018
& 0.876 & 67.11 & 0.319
&0.316  & 73.73 & 0.557 \\

MSBDN~\cite{dong2020multi} & CVPR 2020
& 1.363 & 66.82 & 0.356
& 0.557 & 72.79 & 0.520 \\

Dehamer~\cite{guo2022image} & CVPR 2022
& 1.895 & 66.70 & 0.365
& 0.667 & 72.73 & 0.496 \\

DAD~\cite{shao2020domain} & CVPR 2020
& 1.130 & 66.93 & 0.254
& 0.484 & 71.56 & 0.413 \\

PSD~\cite{chen2021psd} & CVPR 2021
& 0.920 & 72.06 & 0.275
& 0.416 & 76.02 & 0.547 \\

D4~\cite{yang2022d4} & CVPR 2022
& 1.358 & 66.84 & 0.340
& 0.411 & 73.13 & 0.561 \\

RIDCP~\cite{wu2023ridcp} & CVPR 2023
& 0.944 & 70.82 & 0.337
& 0.408 & 74.64 & 0.470 \\

CORUN~\cite{fang2024real} & NeurIPS 2024
& 0.824 & 72.95 & \textbf{\color{second}{0.450}}
& 0.338 & 76.28 & \textbf{\color{second}{0.564}} \\

CoA~\cite{ma2025coa} & CVPR 2025
& 0.859 & 70.38 & 0.349
& 0.314 & 74.27 & 0.536 \\

HazeFlow~\cite{shin2025hazeflow} & ICCV 2025
& \textbf{\color{second}{0.583}} & \textbf{\color{second}{72.96}} & 0.437
& \textbf{\color{second}{0.264}} & \textbf{\color{second}{76.44}} & 0.549 \\
\midrule
PRISM & ---
& \textbf{\color{best}{0.470}} & \textbf{\color{best}{74.05}} & \textbf{\color{best}{0.456}}
& \textbf{\color{best}{0.225}} & \textbf{\color{best}{76.86}} & \textbf{\color{best}{0.590}} \\
\bottomrule
\end{tabular}}
\vspace{-2mm}
\end{table*}

\subsection{Experimental Setup}

We evaluate PRISM on real-world dehazing benchmarks, using unlabeled real hazy images for adaptation and standard test sets for evaluation. Experiments are mainly conducted on RTTS~\cite{li2019benchmarking}, with cross-dataset results on Fattal's dataset~\cite{fattal2014dehazing}. Details on data, metrics, additional results, training settings, supplementary experiments, and visualizations are provided in \Cref{sec:appendix_exp_info}.

\subsection{Comparative Evaluation}

\begin{figure*}[t]
	\centering
	\setlength{\abovecaptionskip}{-0.2cm}
	\begin{center}
		\includegraphics[width=\linewidth]{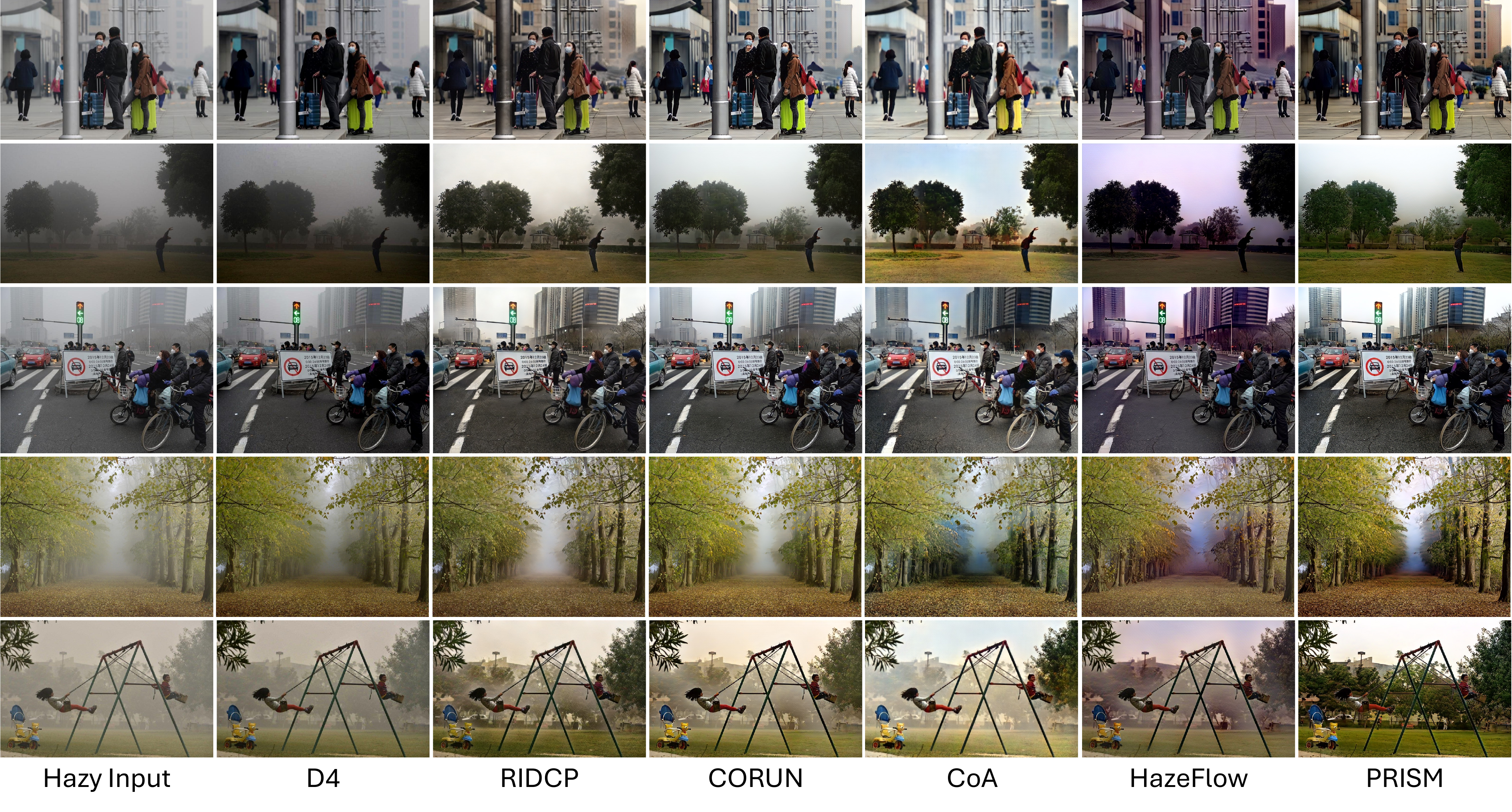}
	\end{center}
 \vspace{-1mm}
	\caption{Visual comparison on the RTTS dataset~\cite{li2019benchmarking} and Fattal's dataset~\cite{fattal2014dehazing}.}
	\label{fig:rtts_fattal}
\vspace{-2mm}
\end{figure*}

\noindent \textbf{Quantitative Comparison.} On the RTTS benchmark (\Cref{tab:rtts_fattal}), PRISM consistently outperforms existing methods, including physics-based models such as CORUN~\cite{fang2024real} and generative methods such as HazeFlow~\cite{shin2025hazeflow}. PRISM achieves the lowest FADE score, indicating its strong ability to remove dense real-world haze, while also obtaining the best results across all adopted non-reference perceptual metrics. On the out-of-distribution Fattal's dataset, PRISM also maintains strong generalization, achieving the best FADE score and leading perceptual quality. These results demonstrate the effectiveness and robustness of PRISM under the adopted real-world dehazing protocol.

\noindent \textbf{Qualitative Comparison.} As shown in \Cref{fig:head,fig:rtts_fattal,fig:more_vis}, PRISM visually outperforms state-of-the-art methods. Baselines like D4~\cite{yang2022d4} and RIDCP~\cite{wu2023ridcp} struggle with dense, non-uniform haze, leaving a noticeable residual veil. While CORUN~\cite{fang2024real} and CoA~\cite{ma2025coa} improve visibility, they often introduce unnatural color shifts and dark artifacts. Similarly, HazeFlow~\cite{shin2025hazeflow} exhibits severe color degradation, hallucinating unrealistic purple hues in background skies. In contrast, PRISM removes substantial haze while preserving more natural colors, sharp textures, and illumination, supporting the effectiveness of our physically constrained adaptation framework.

\subsection{Ablation Study}

\begin{table*}[t]
\centering
\renewcommand{\arraystretch}{0.92}

\begin{minipage}{1.0\linewidth}
\centering
\caption{Ablation studies on data pipeline and PSAR.}
\label{tab:psar_abl}
\setlength{\tabcolsep}{2.8mm}
\resizebox{1.0\linewidth}{!}{
\begin{tabular}{l|l|c|ccc|c|cc}
\toprule
Datasets & Metrics &
\cellcolor{c2!50} w/o non-uni &
\cellcolor{c2!50} w/o $f^{k}_T$ & \cellcolor{c2!50} w/o $f^{k}_J$ &
\cellcolor{c2!50} w/ $f^{k}_A$ &
\cellcolor{c2!50} w/o $spatial~A$ & \cellcolor{c2!50} PSAR  & \cellcolor{c2!50} PRISM \\
\midrule
\multirow{3}{*}{RTTS} & FADE$\downarrow$ &0.955  &0.875  &1.404  &0.885 &0.927 &0.869 &0.470\\
& PAQ2PIQ$\uparrow$      &69.84  &71.01  &60.92  &70.56 &70.31 &70.43 &74.05 \\
& CLIPIQA$\uparrow$ &0.395 &0.380 &0.259 &0.356 &0.381 &0.436 &0.456 \\
\bottomrule
\end{tabular}}
\end{minipage}

\begin{minipage}{1.0\linewidth}
\centering
\caption{Ablation studies on PSAR stage and SSDA.}
\label{tab:psar_abl2}
\setlength{\tabcolsep}{3mm}
\resizebox{1.0\linewidth}{!}{
\begin{tabular}{l|l|ccc|c|cc|c}
\toprule
Datasets & Metrics  & 
\cellcolor{c2!50} $1~Stage$ & \cellcolor{c2!50} $2~Stage$ &
\cellcolor{c2!50} $6~Stage$ &
\cellcolor{c2!50} PSAR &
\cellcolor{c2!50} w/o QGSD &
\cellcolor{c2!50} w/o PGSR & \cellcolor{c2!50} PRISM \\
\midrule
\multirow{3}{*}{RTTS} & FADE$\downarrow$    &0.998  &0.931  &0.840  &0.869  &0.549 &0.652  &0.470\\
& PAQ2PIQ$\uparrow$  &70.11  &70.52  &70.68  &70.43  &73.19  &73.04 &74.05  \\
& CLIPIQA$\uparrow$ &0.355 &0.408 &0.370 &0.436 &0.442 &0.463 &0.456 \\
\bottomrule
\end{tabular}}
\end{minipage}

\begin{minipage}{1.0\linewidth}
\centering
\caption{Generalization studies of SSDA.}
\label{tab:psar_abl3}
\setlength{\tabcolsep}{3.1mm}
\resizebox{1.0\linewidth}{!}{
\begin{tabular}{l|l|cc|cc|cc}
\toprule
Datasets & Metrics  & 
\cellcolor{c2!50} DGUNet~\cite{mou2022deep} & \cellcolor{c2!50} with QGSD &
\cellcolor{c2!50} CORUN$_{s1}$~\cite{fang2024real} &
\cellcolor{c2!50} with SSDA 
& \cellcolor{c2!50}PSAR & \cellcolor{c2!50}PRISM \\
\midrule
\multirow{3}{*}{RTTS} 
& FADE$\downarrow$  &1.111   &0.940  &1.044  &0.745  &0.869 &0.470 \\
& PAQ2PIQ$\uparrow$  &69.92   &72.48  &70.62  &73.56  &70.43 &74.05 \\
& CLIPIQA$\uparrow$ &0.360 &0.375 &0.419 &0.453 &0.436 &0.456 \\
\bottomrule
\end{tabular}}
\end{minipage}

\vspace{-4mm}
\end{table*}

\noindent \textbf{Ablations on PSAR modules.}
As reported in \Cref{tab:psar_abl}, removing $f^{k}_J$ degrades all metrics, confirming the importance of scene-radiance refinement. Removing $f^{k}_T$ slightly improves PAQ2PIQ but worsens FADE and CLIPIQA, so we keep it for a better overall trade-off. Adding $f^{k}_A$ degrades FADE and CLIPIQA, supporting our closed-form proximal $A$ update. Replacing spatial $A$ with a global constant consistently degrades all metrics, confirming the need for local atmospheric-light estimation.

\noindent \textbf{Ablations on haze synthesis.}
As shown in \Cref{tab:psar_abl}, replacing our online non-uniform haze synthesis with a standard uniform synthesis pipeline (w/o non-uni) leads to clear performance degradation. This suggests that exposing the model to heterogeneous density and depth-aware variations during pretraining helps it capture complex real-world haze patterns.

\noindent \textbf{Ablations on stage number.}
The stage number in a deep unfolding network controls the trade-off between computation and reconstruction quality. We evaluate PSAR with $k\in\{1,2,4,6\}$. As reported in \Cref{tab:psar_abl2}, 4 stages already provide strong dehazing quality. Further increasing $k$ does not bring consistent gains in dehazing and perceptual quality, while it increases model depth and optimization difficulty, which may slow convergence and accumulate stage-wise errors.

\noindent \textbf{Effect and generalization of SSDA.}
As reported in \Cref{tab:psar_abl2}, both QGSD and PGSR improve PRISM adaptation, and removing either one weakens the balance between haze removal and perceptual quality. Without PGSR, CLIPIQA slightly increases, suggesting improved texture naturalness, but FADE worsens notably, indicating weaker haze removal. This reflects the trade-off between perceptual texture quality and physical haze removal. To assess transferability, we apply SSDA to other models in \Cref{tab:psar_abl3}. For DGUNet~\cite{mou2022deep}, which lacks explicit scattering variables, QGSD alone still brings clear gains. For the physically grounded CORUN~\cite{fang2024real}, full SSDA improves all metrics. These results show that SSDA generalizes across architectures, with stronger gains when physical variables are available.

\section{Conclusion}
\label{sec:conclusion}
In this paper, we propose PRISM, a unified understanding and restoration framework for real-world image dehazing. In PRISM, we introduce Proximal Scattering Atmosphere Reconstruction (PSAR) to jointly optimize the clear scene and scattering variables via an unfolded proximal process. To bridge the synthetic-to-real domain gap, we design an online non-uniform haze synthesis pipeline and a Selective Self-Distillation Adaptation (SSDA) scheme. This enables quality-gated real-image adaptation while using estimated scattering variables to regularize residual haze removal. Experiments show that PRISM is competitive under the adopted real-world dehazing protocol, delivering dehazed results with fewer artifacts in complex scenes.

\clearpage
\bibliographystyle{unsrt}
\bibliography{main}


\clearpage
\appendix

\section{Detailed Derivations of the Proximal Updates}
\label{sec:derivations}

In this section, we provide the derivations for the closed-form proximal updates used in PSAR. At stage $k$, the network first computes the atmospheric light and transmission proximal points, keeps the atmospheric light branch purely proximal with $A_k=\bar{A}_k$, refines the transmission as $T_k=\bar{T}_k + f_T^{(k)}(\bar{T}_k, J_{k-1})$, and then computes the clear-scene proximal point under the refined transmission $T_k$ and atmospheric light $A_k$.

Since the scattering data term is defined pixel-wise and the quadratic proximal penalties are also separable across pixels, all three subproblems can be solved independently at each pixel. Therefore, it is sufficient to derive the update at a generic pixel. For clarity, we omit the spatial index $(x)$ in this appendix unless needed. Thus, throughout this section, $P,J,A\in\mathbb{R}^3$ denote RGB vectors at one pixel, and $T\in\mathbb{R}$ denotes the corresponding scalar transmission. Because $\lambda_A,\lambda_T,\lambda_J>0$, each subproblem is a strictly convex quadratic problem with a unique minimizer, which is obtained by setting the first-order derivative or gradient to zero.

\subsection{Closed-Form Solution for the Atmospheric Light Update $\bar{A}_k$}

With $J_{k-1}$ and $T_{k-1}$ fixed, the atmospheric light proximal subproblem at one pixel is
\begin{equation}
    E(A)
    =
    \frac{1}{2}
    \left\|
        J_{k-1}T_{k-1} + (1-T_{k-1})A - P
    \right\|_2^2
    +
    \frac{\lambda_A}{2}
    \|A-A_{k-1}\|_2^2.
\end{equation}
Here the optimization variable is the RGB vector $A$, while $J_{k-1}$, $T_{k-1}$, $P$, and $A_{k-1}$ are constants. Differentiating with respect to $A$ gives
\begin{equation}
    \frac{\partial E}{\partial A}
    =
    (1-T_{k-1})
    \left(
        J_{k-1}T_{k-1} + (1-T_{k-1})A - P
    \right)
    +
    \lambda_A(A-A_{k-1}).
\end{equation}
Setting the gradient to $\mathbf{0}$ yields
\begin{equation}
    (1-T_{k-1})
    \left(
        J_{k-1}T_{k-1} + (1-T_{k-1})A - P
    \right)
    +
    \lambda_A(A-A_{k-1})
    =
    \mathbf{0}.
\end{equation}
Collecting the terms involving $A$, we obtain
\begin{equation}
    ((1-T_{k-1})^2+\lambda_A)A
    =
    (1-T_{k-1})(P-J_{k-1}T_{k-1})
    +
    \lambda_A A_{k-1}.
\end{equation}
Therefore, the unique minimizer is
\begin{equation}
    \bar{A}_k
    =
    \frac{
        (1-T_{k-1})(P-J_{k-1}T_{k-1})
        +
        \lambda_A A_{k-1}
    }{
        (1-T_{k-1})^2+\lambda_A
    }.
\end{equation}

\subsection{Closed-Form Solution for the Transmission Update $\bar{T}_k$}

After obtaining $A_k=\bar{A}_k$, the transmission proximal subproblem is
\begin{equation}
    E(T)
    =
    \frac{1}{2}
    \left\|
        TJ_{k-1} + (1-T)A_k - P
    \right\|_2^2
    +
    \frac{\lambda_T}{2}
    (T-T_{k-1})^2.
\end{equation}
Here the optimization variable is the scalar $T$, while $J_{k-1}$, $A_k$, $P$, and $T_{k-1}$ are constants. To make the dependence on $T$ explicit, we rewrite the residual as
\begin{equation}
    TJ_{k-1} + (1-T)A_k - P
    =
    T(J_{k-1}-A_k) + A_k - P.
\end{equation}
Substituting this form into the objective and differentiating with respect to $T$ gives
\begin{equation}
    \frac{\partial E}{\partial T}
    =
    \left\langle
        J_{k-1}-A_k,\,
        T(J_{k-1}-A_k)+A_k-P
    \right\rangle
    +
    \lambda_T(T-T_{k-1}).
\end{equation}
Setting the derivative to zero yields
\begin{equation}
    \left\langle
        J_{k-1}-A_k,\,
        T(J_{k-1}-A_k)+A_k-P
    \right\rangle
    +
    \lambda_T(T-T_{k-1})
    =
    0.
\end{equation}
Expanding the inner product gives
\begin{equation}
    T\|J_{k-1}-A_k\|_2^2
    +
    \left\langle
        J_{k-1}-A_k,\,
        A_k-P
    \right\rangle
    +
    \lambda_T T
    -
    \lambda_T T_{k-1}
    =
    0.
\end{equation}
To match the form used in the main text, we rewrite the expression using $A_k-J_{k-1}$. Since
\[
J_{k-1}-A_k = -(A_k-J_{k-1}),
\]
we have
\[
\|J_{k-1}-A_k\|_2^2 = \|A_k-J_{k-1}\|_2^2
\]
and
\[
\left\langle
J_{k-1}-A_k,\,
A_k-P
\right\rangle
=
-
\left\langle
A_k-J_{k-1},\,
A_k-P
\right\rangle.
\]
Hence, the optimality condition becomes
\begin{equation}
    T\|A_k-J_{k-1}\|_2^2
    -
    \left\langle
        A_k-J_{k-1},\,
        A_k-P
    \right\rangle
    +
    \lambda_T T
    -
    \lambda_T T_{k-1}
    =
    0.
\end{equation}
Collecting the terms involving $T$, we obtain
\begin{equation}
    T
    \left(
        \lambda_T + \|A_k-J_{k-1}\|_2^2
    \right)
    =
    \lambda_T T_{k-1}
    +
    \left\langle
        A_k-J_{k-1},\,
        A_k-P
    \right\rangle.
\end{equation}
Therefore,
\begin{equation}
    \bar{T}_k
    =
    \frac{
        \lambda_T T_{k-1}
        +
        \left\langle
            A_k-J_{k-1},\,
            A_k-P
        \right\rangle
    }{
        \lambda_T + \|A_k-J_{k-1}\|_2^2
    }.
\end{equation}
Expanding the inner product and squared norm over the three RGB channels gives
\begin{equation}
    \bar{T}_k
    =
    \frac{
        \lambda_T T_{k-1}
        +
        \displaystyle\sum_{c=1}^3
        (A_k^{(c)}-J_{k-1}^{(c)})
        (A_k^{(c)}-P^{(c)})
    }{
        \lambda_T
        +
        \displaystyle\sum_{c=1}^3
        (A_k^{(c)}-J_{k-1}^{(c)})^2
    }.
\end{equation}

\subsection{Closed-Form Solution for the Clear-Scene Update $\bar{J}_k$}

After the transmission proximal update, the network refines the transmission as
\begin{equation}
    T_k = \bar{T}_k + f_T^{(k)}(\bar{T}_k,J_{k-1}).
\end{equation}
The clear-scene proximal step is then performed under the refined transmission $T_k$ and the atmospheric light $A_k$. At one pixel, the corresponding subproblem is
\begin{equation}
    E(J)
    =
    \frac{1}{2}
    \left\|
        T_k J + (1-T_k)A_k - P
    \right\|_2^2
    +
    \frac{\lambda_J}{2}
    \|J-J_{k-1}\|_2^2.
\end{equation}
Here the optimization variable is the RGB vector $J$, while $T_k$, $A_k$, $P$, and $J_{k-1}$ are constants. Differentiating with respect to $J$ gives
\begin{equation}
    \frac{\partial E}{\partial J}
    =
    T_k
    \left(
        T_k J + (1-T_k)A_k - P
    \right)
    +
    \lambda_J(J-J_{k-1}).
\end{equation}
Setting the gradient to $\mathbf{0}$ yields
\begin{equation}
    T_k
    \left(
        T_k J + (1-T_k)A_k - P
    \right)
    +
    \lambda_J(J-J_{k-1})
    =
    \mathbf{0}.
\end{equation}
Collecting the terms involving $J$, we obtain
\begin{equation}
    T_k^2 J
    +
    \lambda_J J
    =
    T_k P
    -
    T_k(1-T_k)A_k
    +
    \lambda_J J_{k-1}.
\end{equation}
Using
\[
-T_k(1-T_k)A_k
=
-T_kA_k + T_k^2A_k,
\]
the above equation becomes
\begin{equation}
    (T_k^2+\lambda_J)J
    =
    T_k P
    +
    T_k^2A_k
    -
    T_kA_k
    +
    \lambda_J J_{k-1}.
\end{equation}
Therefore, the unique minimizer is
\begin{equation}
    \bar{J}_k
    =
    \frac{
        T_k P
        + T_k^2A_k
        - T_kA_k
        + \lambda_J J_{k-1}
    }{
        T_k^2+\lambda_J
    }.
\end{equation}
Since the denominator is scalar, it divides each RGB component of the numerator.

\subsection{Summary of the Stage-Wise Variable Flow}

Putting the above derivations together, one PSAR stage follows the sequence
\begin{equation}
\begin{cases}
\bar{A}_k = \Phi_A(P,J_{k-1},T_{k-1},A_{k-1}),\quad A_k=\bar{A}_k,\\[4pt]
\bar{T}_k = \Phi_T(P,J_{k-1},T_{k-1},A_k),\quad
T_k=\bar{T}_k + f_T^{(k)}(\bar{T}_k,J_{k-1}),\\[4pt]
\bar{J}_k = \Phi_J(P,J_{k-1},T_k,A_k),\quad
J_k=\bar{J}_k + f_J^{(k)}(\bar{J}_k,T_k,A_k),
\end{cases}
\end{equation}
where $\Phi_A$, $\Phi_T$, and $\Phi_J$ denote the closed-form proximal maps derived above. Therefore, the appendix derivations are fully consistent with the stage-wise computation described in the main text.

\section{Experiments}

\subsection{Experiment Setup} \label{sec:appendix_exp_info}

\textbf{Data Preparation.} 
We perform initial pre-training of PSAR using the RIDCP dataset~\cite{wu2023ridcp} with the same protocol as HazeFlow~\cite{shin2025hazeflow}. For the adaptation phase, the RESIDE-URHI~\cite{li2019benchmarking} subset, containing 4,807 unpaired real-world hazy images, is used to generate pseudo-labels and refine the network parameters. We evaluate PRISM on the RTTS dataset~\cite{li2019benchmarking}, which contains over 4,000 images characterized by varying resolutions and complex degradation patterns. Additionally, we report results from Fattal's dataset~\cite{fattal2014dehazing} as cross-dataset evidence. To further assess PSAR under full-reference evaluation, we report results on the paired real-world NH-HAZE dataset~\cite{ancuti2020nh}, which contains hazy and corresponding clear image pairs captured under non-homogeneous haze conditions.

\noindent \textbf{Metrics.} 
To assess the performance of the proposed method, we report the results of FADE~\cite{choi2015fade}, PAQ2PIQ~\cite{ying2020patches}, and CLIPIQA~\cite{wang2023exploring}. FADE measures residual haze density, while PAQ2PIQ and CLIPIQA evaluate perceptual quality and naturalness. For paired real-world evaluation on NH-HAZE~\cite{ancuti2020nh}, we further report PSNR and SSIM to measure pixel-level fidelity and structural similarity against the haze-free reference images. To further support the comparison, we provide additional results in \Cref{sec:more_metrics} with more widely used metrics.

\noindent \textbf{Implementation Details.} 
Our framework is trained on four NVIDIA A100 GPUs. The PSAR network contains four unfolding stages. We pretrain the model for 35K iterations using AdamW. The generator and discriminator learning rates are initialized to $2\times10^{-4}$ and $4\times10^{-4}$, respectively, and are halved after 20K iterations. During adaptation, the learning rate is fixed at $2\times10^{-5}$ for 1.5K fine-tuning iterations. We set $\lambda_A^{\text{tv}}=\lambda_c=\lambda_{adv}=0.5$. For efficiency, each \( f_T^{(k)} \) and \( f_J^{(k)} \) is implemented by 3 UNet-style layers with \([1,1,1]\) blocks and 30 input channels, where channels are doubled and spatial resolution is halved with depth.

\subsection{Evaluate PSAR on the Paired Real-world Dehazing Dataset} 
\label{sec:paired}

To further validate the effectiveness of PSAR, we evaluate its performance on the paired real-world dehazing dataset NH-HAZE~\cite{ancuti2020nh}. 

\begin{table*}[h]
\centering
\setlength{\abovecaptionskip}{0cm}
\caption{Quantitative results on the paired real-world dataset NH-HAZE~\cite{ancuti2020nh}.}
\label{tab:paired_datasets}
\setlength{\tabcolsep}{1.3mm}
\resizebox{\textwidth}{!}{
\begin{tabular}{ll|ccccc}
\toprule
Dataset & Metric
& {\cellcolor{gray!20}AOD-Net~\cite{li2017aod}}
& {\cellcolor{gray!20}MSBDN~\cite{dong2020multi}}
& {\cellcolor{gray!20}GCANet~\cite{das2022gca}}
& {\cellcolor{gray!20}FFANet~\cite{qin2020ffa}}
& {\cellcolor{gray!20}DehazeFormer~\cite{song2023vision}} \\
\midrule
\multirow{2}{*}{NH-HAZE~\cite{ancuti2020nh}}
& PSNR$\uparrow$ & 15.31 & 17.34 & 16.64 & 18.48 & 18.15 \\
& SSIM$\uparrow$ & 0.458 & 0.556 & 0.558 & 0.618 & 0.607 \\
\midrule
Dataset & Metric
& {\cellcolor{gray!20}DehazeNet~\cite{cai2016dehazenet}}
& {\cellcolor{gray!20}GridDehazeNet~\cite{liu2019griddehazenet}}
& {\cellcolor{gray!20}UHD~\cite{zheng2021ultra}}
& {\cellcolor{gray!20}DehazeDDPM~\cite{yu2024highqualityimagedehazingdiffusion}}
& {\cellcolor{gray!20}PSAR} \\
\midrule
\multirow{2}{*}{NH-HAZE~\cite{ancuti2020nh}}
& PSNR$\uparrow$ 
& 11.76  & 17.23 & 16.05 & 19.44  & 20.10 \\
& SSIM$\uparrow$ 
& 0.399 & 0.504 & 0.461 & 0.627  & 0.688 \\
\bottomrule
\end{tabular}}
\vspace{-1mm}
\end{table*}

NH-HAZE contains paired hazy and corresponding clear images captured under real-world conditions, providing a direct evaluation of dehazing performance. We report the results in \Cref{tab:paired_datasets}, where PSAR achieves competitive performance compared to state-of-the-art methods, demonstrating its ability to remove haze while preserving image details in real-world scenarios.

\subsection{More Comparative Results} \label{sec:more_metrics}

To reduce metric-specific bias and support a more diverse comparison, we further report RTTS~\cite{li2019benchmarking} results with four additional no-reference image quality assessment (NR-IQA) metrics, complementing the main evaluation with FADE~\cite{choi2015fade}, PAQ2PIQ~\cite{ying2020patches}, and CLIPIQA~\cite{wang2023exploring}: NIMA~\cite{talebi2018nima}, MUSIQ~\cite{ke2021musiq}, ManIQA~\cite{yang2022maniqa}, and LIQE~\cite{zhang2023blind}.
These metrics cover complementary quality aspects: NIMA focuses more on aesthetic quality, while MUSIQ, ManIQA, and LIQE focus more on general perceptual quality.
Together, they provide an additional check on whether the observed improvement remains consistent beyond the main evaluation protocol.

\begin{table*}[h]
\centering
\setlength{\abovecaptionskip}{0cm}
\caption{Supplementary quantitative results on RTTS~\cite{li2019benchmarking} dataset using additional IQA metrics.}
\label{tab:rtts_supp_iqa}
\setlength{\tabcolsep}{3.5mm}
\resizebox{\textwidth}{!}{
\begin{tabular}{l|cccccc}
\toprule
Metric 
& {\cellcolor{gray!20}Hazy Inputs} 
& {\cellcolor{gray!20}PDN~\cite{yang2018proximal}} 
& {\cellcolor{gray!20}MSBDN~\cite{dong2020multi}} 
& {\cellcolor{gray!20}Dehamer~\cite{guo2022image}} 
& {\cellcolor{gray!20}DAD~\cite{shao2020domain}} 
& {\cellcolor{gray!20}PSD~\cite{chen2021psd}} \\
\midrule
NIMA$\uparrow$ 
& 4.857 & 4.880 & 4.896 & 4.898 & 4.778 & 4.783 \\
MUSIQ$\uparrow$ 
& 53.77 & 52.97 & 54.04 & 53.78 & 49.92 & 56.52 \\
\midrule
ManIQA$\uparrow$ 
& 0.311 & 0.280 & 0.306 & 0.316 & 0.225 & 0.296 \\
LIQE$\uparrow$ 
& 1.982 & 1.959 & 2.108 & 2.037 & 1.724 & 1.693 \\
\midrule
Metric 
& {\cellcolor{gray!20}D4~\cite{yang2022d4}} 
& {\cellcolor{gray!20}RIDCP~\cite{wu2023ridcp}} 
& {\cellcolor{gray!20}CORUN~\cite{fang2024real}} 
& {\cellcolor{gray!20}CoA~\cite{ma2025coa}} 
& {\cellcolor{gray!20}HazeFlow~\cite{shin2025hazeflow}} 
& {\cellcolor{gray!20}PRISM} \\
\midrule
NIMA$\uparrow$ 
& 4.852 & 5.241 & 5.301 & 5.059 & 5.296 & 5.347 \\
MUSIQ$\uparrow$ 
& 53.57 & 59.38 & 67.60 & 53.43 & 63.92 & 70.02 \\
\midrule
ManIQA$\uparrow$ 
& 0.300 & 0.275 & 0.357 & 0.243 & 0.346 & 0.362 \\
LIQE$\uparrow$ 
& 2.089 & 2.671 & 3.049 & 1.971 & 2.781 & 3.569 \\
\bottomrule
\end{tabular}}
\vspace{-1mm}
\end{table*}

Since prior dehazing studies often do not specify the NIMA and LIQE evaluation implementations used for evaluation, for NIMA, we consistently adopt the VGG16-AVA version and retest all methods under the same implementation to ensure a fair and reproducible comparison. For LIQE, we only evaluate images whose shorter side is no smaller than 224 pixels, as required by the metric, instead of resizing the inputs. This avoids potential errors from different resizing kernels and the sensitivity of LIQE to resizing.

As shown in~\Cref{tab:rtts_supp_iqa}, PRISM achieves the best results across all four additional NR-IQA metrics on RTTS.
This trend is consistent with the main evaluation using FADE, PAQ2PIQ, and CLIPIQA, indicating that the advantage of PRISM is not tied to a specific metric choice.
The results provide complementary evidence that PRISM produces visually higher-quality dehazed images under different quality assessment models.

\begin{table*}[h]
\centering
\caption{Additional ablations on atmospheric-light rendering and EMA teacher update.}
\label{tab:psar_abl4}
\setlength{\tabcolsep}{1.5mm}
\resizebox{1.0\linewidth}{!}{
\begin{tabular}{l|l|ccc|c|cc|c}
\toprule
Datasets & Metrics  & 
\cellcolor{c2!50} S.R. $A_{\textrm{rend}}$ & \cellcolor{c2!50} S.R. $A_{\textrm{rend}}+\ell_1$ &
\cellcolor{c2!50} $A_{\textrm{rend}}+\ell_1$ &
\cellcolor{c2!50} PSAR &
\cellcolor{c2!50} Fixed Teacher &
\cellcolor{c2!50} w/o EMA & \cellcolor{c2!50} PRISM \\
\midrule
\multirow{3}{*}{RTTS} 
& FADE$\downarrow$  &0.918    &0.920      &0.894    &0.869     &0.583  &0.526  &0.470\\
& PAQ2PIQ$\uparrow$ &69.63    &69.11      &70.18    &70.43     &72.67  &73.12  &74.05\\
&CLIPIQA$\uparrow$  &0.388    &0.376      &0.396    &0.436     &0.438  &0.447  &0.456 \\
\bottomrule
\end{tabular}}
\end{table*}

\subsection{Ablation on EMA Update}
\label{subsec:ema_ablation}
This ablation is designed to check whether the EMA loop improves adaptation or merely reinforces its own pseudo-labels.
In self-distillation, low-quality teacher predictions can be propagated through the teacher update if the teacher and student become too tightly coupled.
We therefore evaluate two teacher-update controls in \Cref{tab:psar_abl4}.
\textit{Fixed Teacher} freezes the teacher after synthetic pretraining, breaking the feedback from the adapted student to the teacher while keeping the same QGSD and PGSR losses.
\textit{w/o EMA} removes temporal weight averaging and uses the online model directly for target generation and physical auditing.
Together, these controls test whether PRISM's EMA teacher provides stable candidate targets under SSDA, rather than improving through unchecked teacher-student coupling.

\subsection{Ablation on Atmospheric Light Offset and Supervision}
\label{subsec:spatial_A_ablation}
We further verify whether applying $\ell_1$ supervision to $A_{\textrm{rend}}$ and further extending the spatial random transformation of $A_{\textrm{rend}}$ can improve performance. The results show that directly supervising $A_{\textrm{rend}}$ does not improve performance and may even cause slight degradation. This may be because direct supervision on the rendered atmospheric light offset restricts the model's representation ability and adaptability. Applying more complex pixel-level spatial random transformations to the rendered atmospheric-light map also brings no performance gain, and may introduce content-irrelevant counterfactual illumination patterns, leading to performance degradation. Therefore, we finally choose not to impose $\ell_1$ supervision on $A_{\textrm{rend}}$ and keep its current random pattern as an atmospheric-light shift to better adapt to real scenes.

\section{Additional Experimental Visualizations}

\subsection{Visualization of Different Training Phases}
\begin{figure*}[!htb]
	\centering
	\setlength{\abovecaptionskip}{0cm}
	\begin{center}
		\includegraphics[width=\linewidth]{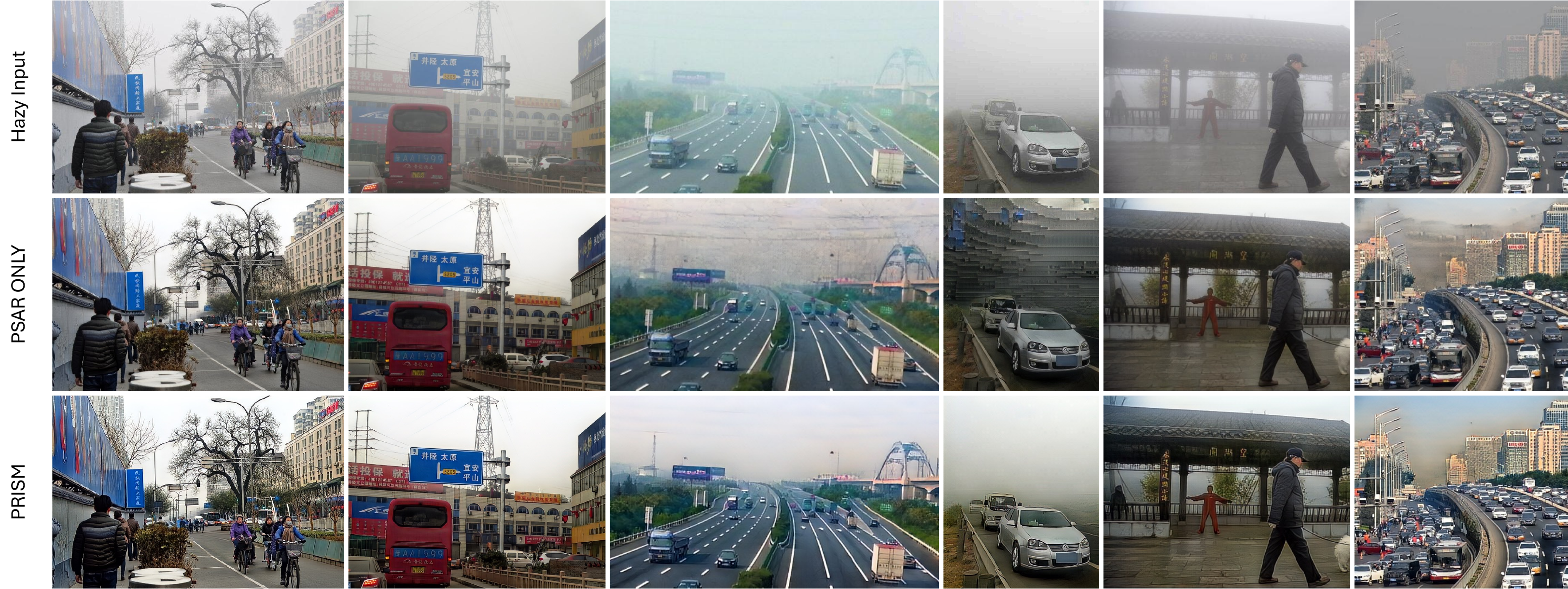}
	\end{center}
	\caption{Visualization of different training phases.}
	\label{fig:training_phases}
\end{figure*}

The \Cref{fig:training_phases} visualizes the results of PSAR at different training phases, including the initial pre-training stage and the subsequent SSDA adaptation stage, namely PRISM. The pre-trained PSAR already removes much of the haze from the input image, but residual haze and color distortion remain, especially in heavily degraded regions. After SSDA adaptation, PRISM produces a cleaner result with fewer haze artifacts, more natural colors, and better detail preservation in dense haze or strong color-cast areas. These results show that SSDA further improves PSAR on real-world hazy images through selective self-distillation and residual haze auditing.

\subsection{Visualization of the Estimated $A$, $T$, and $J$}

\begin{figure*}[!htb]
	\centering
	\setlength{\abovecaptionskip}{0cm}
	\begin{center}
		\includegraphics[width=\linewidth]{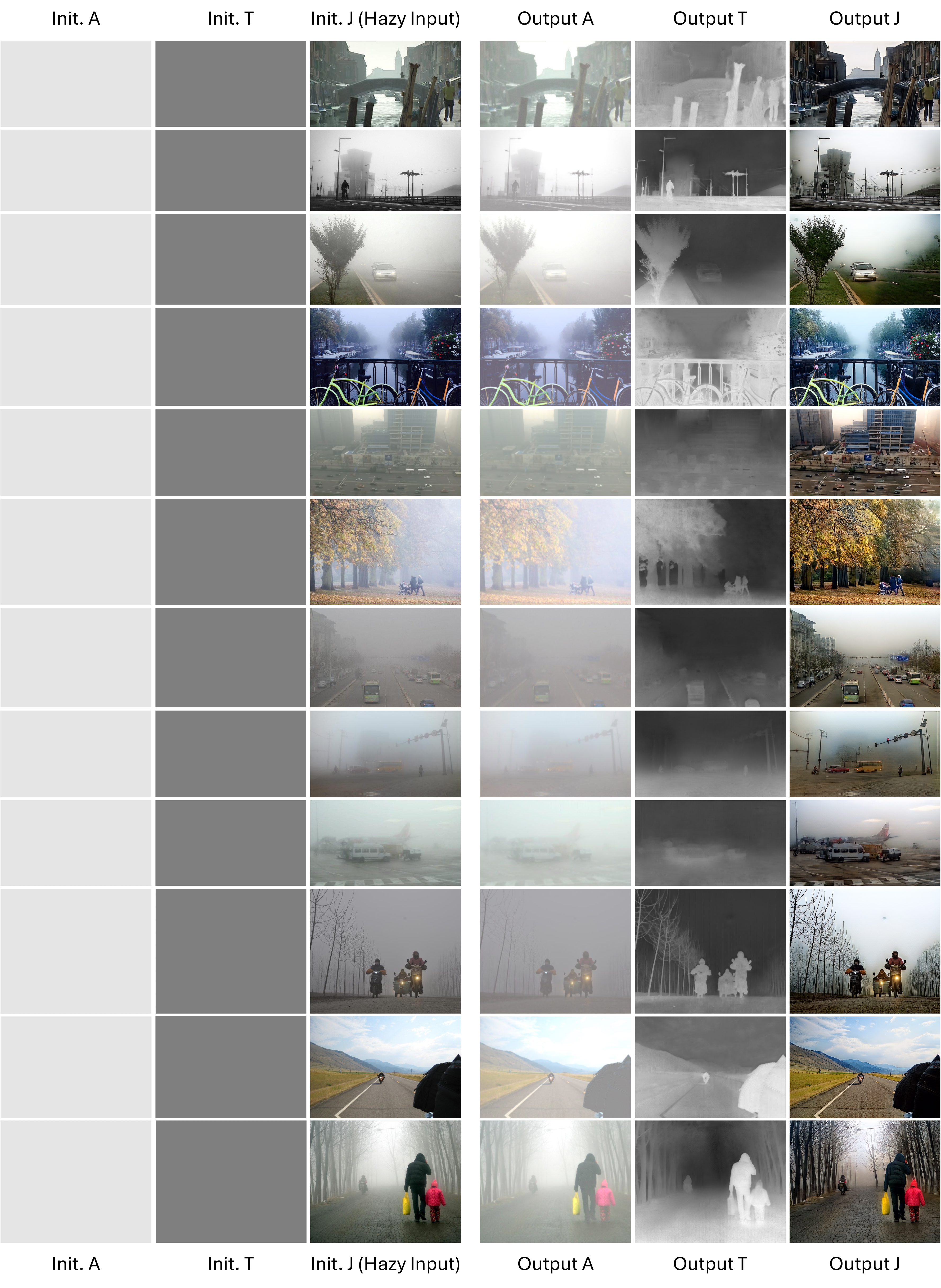}
	\end{center}
	\caption{Visualization of the initialized inputs and the decomposed components estimated by the proposed PRISM, including atmospheric light $A$, transmission $T$, and restored scene radiance $J$.}
	\label{fig:process}
\end{figure*}

In \Cref{fig:process}, we visualize the inputs and outputs of PSAR. For the inputs, we initialize $J$ with the hazy image $P$, set the atmospheric light $A$ to $0.9$, and initialize the transmission map $T$ to $0.5$. The outputs $A$, $T$, and $J$ correspond to the atmospheric light map, transmission map, and dehazed image estimated by PSAR from these initial states.

The visualization shows that PSAR produces transmission maps with clear spatial variation, indicating that it can capture non-uniform haze concentration across different regions of the scene. Regions with heavier haze usually correspond to lower transmission values, while clearer regions tend to have higher transmission. Meanwhile, the estimated atmospheric light maps reflect scene-dependent illumination and color bias rather than collapsing to a trivial constant pattern. This helps separate scattering-related illumination effects from the underlying scene radiance. As a result, PSAR produces clean and visually faithful dehazed images with more natural colors and fewer visible degradation artifacts.
These results suggest that PSAR learns a meaningful decomposition of the hazy observation into atmospheric light, transmission, and scene radiance, instead of merely enhancing local contrast.

\subsection{More Visualization Results}

\begin{figure*}[!htb]
	\centering
	\setlength{\abovecaptionskip}{0cm}
	\begin{center}
		\includegraphics[width=\linewidth]{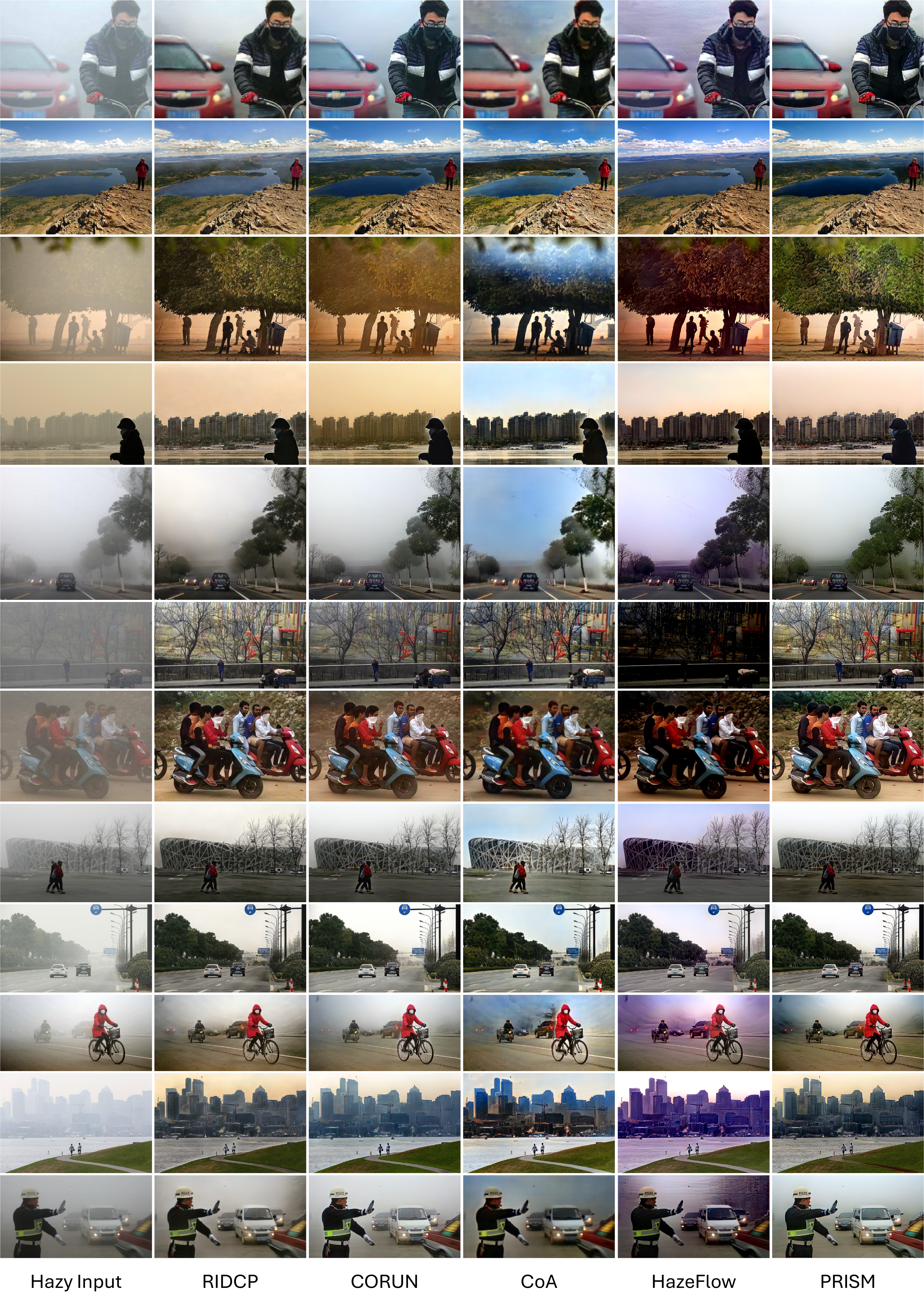}
	\end{center}
	\caption{More visualization results of our PRISM and other state-of-the-art methods.}
	\label{fig:more_vis}
\end{figure*}

We provide more visual comparisons with RIDCP~\cite{wu2023ridcp}, CORUN~\cite{fang2024real}, CoA~\cite{ma2025coa}, and HazeFlow~\cite{shin2025hazeflow} in \Cref{fig:more_vis}. Compared with these methods, PSAR recovers cleaner scene structures, more natural contrast, and more faithful colors under challenging real-world non-uniform haze. In particular, RIDCP and CORUN may still leave noticeable residual haze in heavily degraded regions, while CoA and HazeFlow can produce local color distortion or over-enhancement in some cases. By contrast, PSAR removes haze more thoroughly while better preserving the original scene appearance.

This advantage is especially clear in regions with dense haze, uneven illumination, or strong color cast, where competing methods often struggle to balance dehazing strength and visual fidelity. Overall, PSAR achieves a better trade-off between haze removal, structure preservation, and color realism, leading to higher-quality dehazed results.

\section{Failure Cases Analysis}

\begin{figure*}[h]
	\centering
	\setlength{\abovecaptionskip}{0cm}
	\begin{center}
		\includegraphics[width=0.8\linewidth]{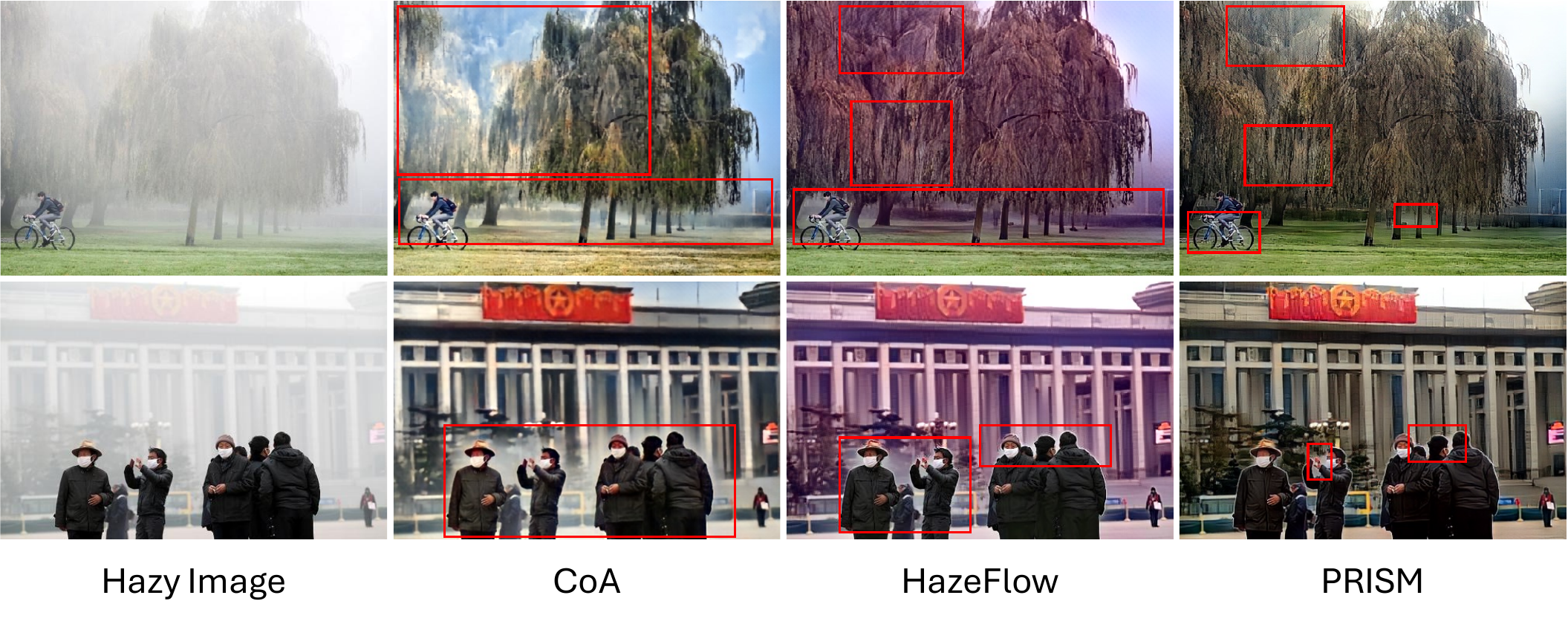}
	\end{center}
	\caption{Failure cases of our proposed PSAR.}
	\label{fig:failure_cases}
\end{figure*}

Representative failure cases are shown in~\Cref{fig:failure_cases}. Although PRISM produces cleaner results than previous methods, extremely dense haze near object boundaries can still lead to mild residual artifacts. These regions are challenging because strong scattering, weak contrast, and depth discontinuities are tightly mixed. Nevertheless, compared with prior methods that often leave broader haze patches or introduce visible distortions, PRISM better suppresses haze while preserving scene structures.

\section{Limitations and Future Work}
\label{sec:limitations}

Although PSAR is highly effective for real-world haze removal, its current formulation is still mainly designed around atmospheric scattering caused by haze~\cite{fang2024real,feng2024advancing}. Therefore, its applicability to special imaging scenarios, such as medical imaging~\cite{fang2026photon}, snow~\cite{valanarasu2022transweather}, rain~\cite{valanarasu2022transweather}, underwater imaging~\cite{chiang2011underwater}, or low-illumination scenes~\cite{ma2022toward,jin2022unsupervised,jin2023enhancing,he2023reti}, has not yet been fully studied. Since SSDA operates at the adaptation level, it may still provide useful guidance beyond standard haze settings. Extending PSAR to a broader range of weather and imaging conditions, while further examining the generalization behavior of SSDA in these scenarios, is an important direction for future work. In addition, although PRISM achieves strong quantitative and visual results, current real-world dehazing evaluation remains incomplete. Existing IQA metrics are useful for benchmarking, but they do not always fully reflect perceptual factors such as color fidelity, halo suppression, and structural faithfulness. Future work will explore more comprehensive haze-aware evaluation protocols and task-driven settings to better assess the practical value of physics-guided dehazing.

\section{Broader Impacts}
\label{sec:broader_impacts}

Real-world image dehazing is an important image restoration task that removes haze degradation from images captured in real-world scenarios, with applications in autonomous driving, remote sensing, security monitoring, and general photography. This paper introduces PRISM, a physically structured dehazing framework that combines online non-uniform haze synthesis with selective self-distillation adaptation, achieving strong performance on real-world dehazing benchmarks. Image dehazing methods aim to restore degraded visual observations rather than generate new semantic content, and have not been associated with direct negative societal impacts.

\clearpage

\end{document}